\documentclass{esannV2}
\usepackage[dvips]{graphicx}
\usepackage[latin1]{inputenc}
\usepackage{amssymb,amsmath,array}
\usepackage{multirow}
\begin{document}
\title{Action-Based ADHD Diagnosis in Video}

\author{Yichun Li$^{1}$, Yuxing Yang$^1$, Rajesh Nair$^2$, Syed Mohsen Naqvi$^1$
%
%
\vspace{.3cm}\\
%
\textsuperscript{\rm 1}Intelligent Sensing and Communications Research Group, Newcastle University, UK.\\
\textsuperscript{\rm 2}Cumbria, Northumberland, Tyne and Wear (CNTW), NHS Foundation Trust, UK
%
}

\maketitle

\begin{abstract}
Attention Deficit Hyperactivity Disorder (ADHD) causes significant impairment in various domains. Early diagnosis of ADHD and treatment could significantly improve the quality of life and functioning. Recently, machine learning methods have improved the accuracy and efficiency of the ADHD diagnosis process. However, the cost of the equipment and trained staff required by the existing methods are generally huge. Therefore, we introduce the video-based frame-level action recognition network to ADHD diagnosis for the first time. We also record a real multi-modal ADHD dataset and extract three action classes from the video modality for ADHD diagnosis.  The whole process data have been reported to CNTW-NHS Foundation Trust, which would be reviewed by medical consultants/professionals and will be made public in due course.
\end{abstract}

\section{Introduction}
Attention deficit hyperactivity disorder (ADHD) is a worldwide prevalent neurodevelopmental disorder. While the adult population has a high rate of undiagnosed and has reached 3\% of the population  \cite{nash2022machine,huang2020conditional}. ADHD patients exhibit inattention, impulsivity, and hyperactivity symptoms, with detrimental effects on brain development \cite{loh2022automated, tenev2014machine}. 


In recent years, machine learning methods and deep learning algorithms have been used in ADHD diagnosis and classification \cite{duda2016use,yang2022two}. Most of the research is based on Magnetic Resonance Imaging (MRI), Electroencephalography (EEG), and natural language processing which achieves high accuracy \cite{luo2020multimodal,li2019identifying,tang2022adhd}, but also with a high cost of equipment and operational staff.  Hence, we propose a new low-cost ADHD diagnosis approach on a machine learning-based ADHD action detection network in this work. We use video because it is easy to capture the action performance of the participants, and it can greatly reduce the cost of diagnosis. The main contributions of our work are listed as follows: 1) an attention test is designed for multi-modal ADHD real data recording. 2) an ADHD diagnosis system based on 3D-CNN action recognition is implemented, and video data is evaluated with different network structures; 3) classification criteria is also proposed to provide diagnosis results with time-action ADHD characteristics.

\section{Participants and Procedure}

We recorded a multi-modal ADHD dataset which includes 7 ADHD subjects diagnosed by the NHS medical consultant under the DSM-V criteria and 10 neurotypical controls. The gender distribution for 7 subjects is 3 males and 4 females, provided by the CNTW-NHS Foundation Trust. The control group consists of 9 males and 1 female. All participants are adults aged between 18 and 50. For the control group, adults who did not have neurological problems and ADHD diagnosis history were the volunteers from Newcastle University.

An attention and responsiveness test is provided for all participants. We prepare four continuous dialogue tasks: 1) a brief conversation between the participants and the interviewer, approximately 10-20 minutes long; 2) performing Cambridge Neuropsychological Test Automated Battery (CANTAB) tasks. This task takes about 40-50 minutes; 3) beep reaction task. This task takes 6 minutes; 4) watching videos, including a math video labelled `boring' and a rally video labelled `exciting'. This task takes 10 minutes. The video signals are recorded by 3 GoPro cameras which contain a front-faced camera 1 to record facial information and two side cameras 2\&3 to record the information of the left and right torsos and limbs with a resolution of $3840\times 2160$.


\begin{figure}[ht!]
\centering
\includegraphics[scale=0.5]{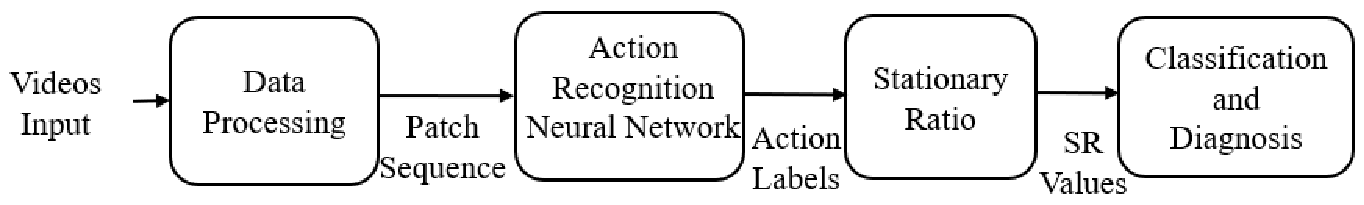}
\caption{Flow diagram of the ADHD diagnosis system. The training dataset for the action recognition function is based on 3 classes and is named ADHD-3. The three action classes are still-position, limb-fridges, and torso-movements. }\label{Fig:MV}
\end{figure}

The block diagram of the proposed ADHD diagnosis system is shown in Fig. 1. The system contains four main parts: data processing, action recognition, stationary ratio calculation, and ADHD diagnosis. Existing action recognition datasets are not focused on typical ADHD symptoms, e.g., fidgeting of the limbs and the body when the subjects and controls are in a sitting position during the data recording. Specifically, the training dataset used in the proposed action recognition module mainly focuses on continuous actions (duration over five seconds) in the sitting position. The ADHD diagnosis result is summarized and classified by estimating the distribution of action labels of the action recognition part with a novel evaluation matrix named stationary ratio (SR).

Since the raw frame size from recorded videos is too large to feed into the diagnosis system, the input frame is reduced from $3840\times 2160$ to $320\times 180$. The landmark of the participant's waist is the center of the processed frame in the sitting position. The video sequences are also down-sampled from 32FPS to 16FPS to reduce the computational cost. Then, after the frame segmentation and patch extraction step, the patches with the size $180\times 180$ containing the samples' torso and limb information are used for training the network.

We propose a novel measurement named Stationary Ratio (SR) as the evaluation criterion for action classification of ADHD symptoms detection. It focuses on the percentage of periods that the test subject is at the still position. The SR is defined as:

\begin{equation}
 SR=\alpha _{1} /(\alpha _{1}+\alpha _{2}+\alpha _{3})
\end{equation}
where $\alpha _{1}$ denotes the number of the samples of predicted still position, $\alpha _{2}$ is the number of samples of small ranges (less than $30^\circ$) of limb fidgets, and $\alpha _{3}$ is the number of the samples of large rotations (more than $30^\circ$) of torso movements.


As aforementioned, we use Camera 2 and Camera 3 for left and right viewpoints, respectively. Therefore, we use the average SR measurement of the left and right viewpoint as $SR_{Avg}$.

\section{Experiments}
\subsection{Datasets and Data Processing}


The action recognition experiments use the three-class action recognition dataset, i.e., still-position, which contains 88 video clips, limb-fidgets with 110 clips, and torso movements with 101 clips. Each of the clips is between 10-15 seconds. The training, validation, and testing data split is 6/2/2, respectively. The diagnosis dataset for the whole system consists of 34 videos, including 7 subjects and 10 controls of the whole process videos from the left and right sides, and the length of each video is 60-90 minutes. Actions are labeled per three frames in the training, testing, and diagnosis steps.


\subsection{Experiment Set up and Comparisons}
We choose a 3D-CNN structure (C3D) as the main core network \cite{tran2015learning,lecun2015deep}. 
There are 8 convolution layers that have 3$\times$3$\times$3 kernels with 1 stride. Different from the original C3D structure, we add a fully connected layer to fit the size of the input data. The probabilities of each action are obtained with three fully connected layers with 8192 units and a Softmax activation.
 
The loss $\mathcal{L}_{c}$ of the training process is to minimize the cross entropy of the outputs and true labels results:
\begin{equation}
L_{c} =-\sum_{i}^{} P_{l}(i)\log_{}{P_{o}(i)}
\end{equation}
where $i$th means the set of labels with $n$ length, $P_{l}$ and $P_{o}$are the distribution of true labels, and the distribution of classification output, respectively.

The training epochs for the action classification are 80, and the learning rate is $1\times 10^{-9}$. All the experiments are run on a workstation with four Nvidia GTX 1080 GPUs and 16 GB of RAM.


\subsection{Action Recognition and ADHD Diagnosis Results}

In these experiments, the SR performance of 7 subjects and 10 controls is evaluated with $SR_{L}$, $SR_{R}$, and $SR_{Avg}$. The results are shown in Table 1. 

\begin{table*}[htbp!]
\centering
\small\addtolength{\tabcolsep}{-5pt}
\caption{The comparisons of the stationary ratio (SR) for the overall subjects and controls. And 'S', 'C', 'F', 'M' indicate subject, controls, female, and male, respectively. Each result is the average of 5 experiments.\\}
\begin{tabular}{c|c c c c c c c c c c}
\hline
 \multirow{2}{*}{\textbf{Samples}} & S2 (M) & S6 (M) & S9 (F) & S10 (F) & S12 (F) & S13 (M) & S14 (F) & C1 (M) & C3 (M) \\ 
                                   & C4 (M) & C5 (M) & C7 (M) & C8 (M) & C11 (M) & C15 (M) & C16 (F) & C17 (M) \\ 
\hline
 \multirow{2}{*}{$\mathbf{SR_{Avg}}$} & 0.39 & 0.24 & 0.64 & 0.36 & 0.53 & 0.75 & 0.66 & 0.86 & 0.91 \\ 
                                     & 0.97 & 0.73 & 0.89 & 0.90 & 0.95 & 0.87 & 0.73 & 0.77 \\ 
\hline
\end{tabular}
\end{table*}

From Table 1, the average of $SR_{Avg}$ for all 17 participants is 0.71. Particularly, the average of $SR_{Avg}$ for 7 subjects and 10 controls are 0.50 and 0.86, respectively. Therefore, 0.71 is adapted as the threshold for the ADHD diagnosis. In the group of subjects, it is highlighted that only Subject 13 has the abnormal $SR_{Avg}$ of 0.75. We have sent requests to the clinicians of CNTW-NHS Foundation Trust to query and double-check the diagnosis details of this ADHD subject. Further analysis will be a future work and meanwhile can be considered as a failure case, in case the clinician will confirm Subject 13. 

Based on the threshold value, i.e., 0.71, we further calculate the precision, sensitivity, accuracy, and the Area Under Curve (AUC) of two traditional neural networks: R2Plus1D and R3D \cite{targ2016resnet, tran2018closer}, and our proposed 3D-CNN framework in Table 2. 

\begin{table}[htbp!]
\centering
\small\addtolength{\tabcolsep}{-5pt}
\caption{ADHD diagnosis system performance with different neural networks. \\}
\begin{tabular}{c|c c c c}
\hline
\multicolumn{1}{l|}{}&\textbf{Sensitivity }(\%) &\textbf{Precision}(\%) &\textbf{Accuracy}(\%)&\textbf{AUC} \\ \hline
R3D \cite{tran2018closer}& 100.0 & 58.8 & 58.8 & 0.50  \\ \hline
R2Plus1D \cite{targ2016resnet}& 100.0 & 66.7 & 70.6 & 0.56  \\ \hline
3D-CNN & \textbf{100.0} & \textbf{90.9} &\textbf{94.1} & \textbf{0.97} \\ \hline
\end{tabular}
\end{table}

From Table 2, the proposed model shows better performance than the R3D and R2Plus1D. Because the proposed method concentrates on the features from both the spatial and the temporal dimensions, thereby capturing the action information encoded in multiple adjacent frames, which plays an important role in ADHD typical human action recognition \cite{ji20123d}. Therefore, the proposed method shows high sensitivity in the recognition results of the small range of limb fidgets and improves the performance of ADHD diagnosis results.

\subsection{Time-Action Based Analysis}

According to DSM-V, some symptoms of hyperactivity-impulsivity are observable in ADHD adults, such as difficulty in sitting still, fidgeting legs, tapping with a pen, etc. \cite{edition2013diagnostic}. However, it is hard to record manually during the traditional diagnostic process. Through our system, the actions of each participant are fully captured and visualized. Fig. 2 shows the timeline bar chart from the classification results of the ADHD subject and control groups. 

\begin{figure}[ht!]
\centering
\includegraphics[scale=0.55]{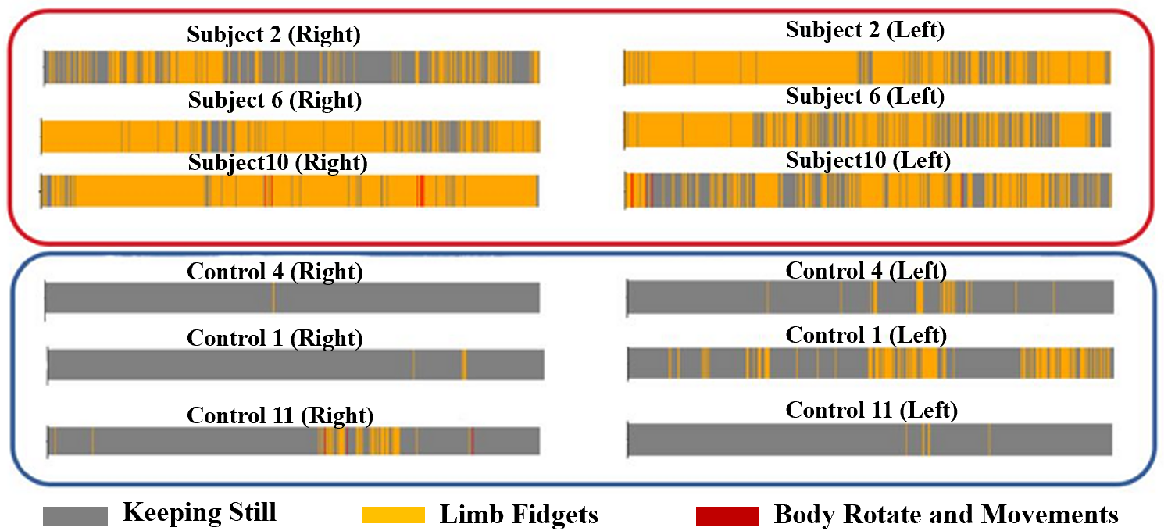}
\caption{Action change timeline chart of three ADHD subjects (top in red box) and three controls (bottom in blue box) recorded by Camera2 (left) and Camera3 (right). }\label{Fig:MV}
\end{figure}

From Fig. 2, the proportion of gray parts (keeping still or almost stationary) in the ADHD subjects group is obviously lower than that in the controls group, which is consistent with clinical observations.

\subsection{Comparison with State-of-the-Art}
Table 3 shows the performance of the state-of-the-art ADHD diagnosis systems on the different datasets containing EEG and trajectory signals collected by wearable sensors.

\begin{table}[htbp!]
\centering
\small\addtolength{\tabcolsep}{-5pt}
\caption{The ADHD diagnosis performance of state-of-the-art methods with the proposed method, where 'S' mean  ADHD subjects, and 'C' mean controls.\\}
\begin{tabular}{c|c|c|c|c}
\hline
Author&S/C& Data Input& Classifier& Accuracy \\ \hline
Luo et al.\cite{luo2020multimodal}& 36/ 36 & MRI \& DTI & CNN & 76.6\% \\ \hline
Dubreuil et al.\cite{dubreuil2020deep}& 20/ 30 & EEG & CNN & 88.0\% \\ \hline
Munoz et al.\cite{munoz2018automatic}& 11/ 11 & Trajectory & CNN & 93.8\% \\ \hline
Proposed method & 7/ 10& Videos&3D-CNN &\textbf{94.1\%} \\ \hline
\end{tabular}
\end{table}
\vspace{-0.9em}
From Table 3, the proposed method outperforms the state-of-the-art ADHD diagnosis methods. Compared to the machine learning methods for ADHD diagnosis, our proposed action-based framework can more intuitively observe ADHD-related action rules. Therefore, the generalization and applicability are improved.

\section{Conclusions}
This paper proposed an ADHD diagnosis system based on the action recognition framework. Meanwhile, a novel measure was proposed to evaluate the action recognition results. The experimental results showed that our system outperformed the state-of-the-art methods regarding precision, accuracy, and AUC. Moreover, the proposed method is less expensive and suitable for a broad range of initial ADHD diagnoses compared with the existing neuroscience diagnostic methods. In our future work, we will extend the dataset to further cover real-world patient distribution and consider recording more multi-modal data, e.g., EEG and fMRI, to perform fusion and evaluate related results.

\begin{footnotesize}
\bibliographystyle{unsrt}
\bibliography{output}

\end{footnotesize}


\end{document}